# CLUSTER EXPANSIONS AND ITERATIVE SCALING
# FOR MAXIMUM ENTROPY LANGUAGE MODELS

JOHN D. LAFFERTY AND BERNHARD SUHM
*School of Computer Science*
*Carnegie Mellon University*
*5000 Forbes Avenue*
*Pittsburgh, PA 15217 USA*[†]

**Abstract.** The maximum entropy method has recently been successfully introduced to a variety of natural language applications. In each of these applications, however, the power of the maximum entropy method is achieved at the cost of a considerable increase in computational requirements. In this paper we present a technique, closely related to the classical cluster expansion from statistical mechanics, for reducing the computational demands necessary to calculate conditional maximum entropy language models.

## 1. Introduction

In this paper we present a computational technique that can enable faster calculation of maximum entropy models. The starting point for our method is an algorithm [1] for constructing maximum entropy distributions that is an extension of the generalized iterative scaling algorithm of Darroch and Ratcliff [2,3]. The extended algorithm relaxes the assumption of [2,3] that the constraint functions sum to a constant, and results in a set of decoupled polynomial equations, one for each feature, that must be solved to obtain the scaling terms. For each iteration, the distribution must be normalized (that is, the partition function must be calculated), and the coefficients of the polynomials must be determined; these steps have roughly the same computational cost.

For language modeling applications the partition function and coefficient calculations entail summing over the target vocabulary, typically on the order of 10,000–100,000 words, and determining those features that apply to each possible word for each context that appears in the training data. When this calculation is implemented directly by carrying out the summation while hashing to determine features and feature weights, it can be exceedingly slow. We address this problem by use of a technique that we call the *cluster expansion*, due to its resemblance to series expansion methods in statistical physics, that carries out both the partition function and coefficient calculations efficiently. Our basic idea is to avoid hashing and an explicit summation over the entire target vocabulary for each context by calculating the partition function (or coefficients) for all contexts simultaneously as a telescoping sum of polynomials in the feature weights. By choosing the data structures in the implementation appropriately, the cluster expansion can be easily implemented for a class of language models that includes $n$-gram constraints in addition to state constraints from an underlying automaton, or other long-distance constraints.

---
[†]Research supported in part by NSF and ARPA under grant IRI-9314969 and the ATR Interpreting Telecommunications Research Laboratories.

In this paper we present a description of the basic technique as well as its application to the construction of a simple language model for use in a speech recognition system.

## 2. Language Modeling

### 2.1. LANGUAGE MODELS AS PRIORS FOR BAYESIAN DECODING

Language modeling attempts to identify regularities in natural language and capture them in a statistical model. Language models are crucial ingredients in automatic speech recognition [4] and statistical machine translation [5] systems, where their use is naturally viewed in terms of the *noisy channel* model from information theory. In this framework an information source emits messages $X$ from a distribution $P(X)$ which then enter into a noisy channel and emerge transformed into observables $Y$ according to a conditional probability distribution $P(Y \mid X)$. The problem of decoding is to determine the message $\widehat{X}$ having the largest posterior probability given the observation:

$$\widehat{X} = \arg\max_{X \in \mathcal{H}} P(X \mid Y) = \arg\max_{X \in \mathcal{H}} P(Y \mid X)\, P(X)\,.$$

Thus, Bayesian decoding is carried out using a prior distribution $P(X)$ on messages, a channel model $P(Y \mid X)$, and a decoder $\arg\max_{X \in \mathcal{H}}$. For speech recognition and machine translation, the prior distribution is called a *language model*, and it must assign a probability to every string of symbols that can be hypothesized by the decoder. The most common language models used in today's speech systems are the $n$-gram models, constructed in terms of simple word frequencies.

### 2.2. CONDITIONAL MAXIMUM ENTROPY LANGUAGE MODELS

In the usual application of the maximum entropy principle [6], prior information, typically in the form of frequencies, is represented as a set of constraints which collectively determine a unique maximum entropy distribution. For example, if we observe certain bigram word frequencies $c_{ij} = \tilde{p}(w_i\, w_j)$ and we constrain a language model to agree with these observations, the maximum entropy distribution assigns a probability $p_\lambda(W)$ to a word string $W$ according to a Gibbs distribution

of the form

$$p_\lambda(W) = \frac{1}{Z_\lambda} \exp\left(\sum_{ij} \lambda_{ij}\, f_{ij}(W)\right)$$

where the *feature* $f_{ij}(W)$ counts the number of times the bigram $w_i w_j$ occurs in the string $W$, and where the partition function $Z_\lambda$ is obtained by summing over all possible word strings $W$.

In contrast to this use of the joint distribution, recent applications of the maximum entropy method in language modeling [7,8] have employed *conditional* models. Such models employ features to represent various frequencies in the training text, such as the bigram features just mentioned, but they use this information to constrain a family of conditional exponential models. Factoring a word string $W = w_0 w_1 \cdots w_N$ into conditional probabilities we can write

$$p(W) = p(w_0) \prod_{i=1}^{N} p(w_i \,|\, w_0 w_1 \cdots w_{i-1}) = p(w_0) \prod_{i=1}^{N} p(w_i \,|\, h_i)$$

where $h_i$ is the *history at time $i$*. In terms of conditional models, the constraints are presented as

$$\sum_h \tilde{p}(h) \sum_w p(w \,|\, h)\, f_\alpha(h, w) = \sum_{h,\,w} \tilde{p}(h, w)\, f_\alpha(h, w)$$

where $h$ is a history, and the maximum entropy model subject to these constraints is given by

$$p_\lambda(w \,|\, h) = \frac{1}{Z_\lambda(h)} \exp\left(\sum_\alpha \lambda_\alpha f_\alpha(h, w)\right). \tag{1}$$

The partition function $Z_\lambda(h)$ is now obtained from summing over the target word vocabulary, rather than over all word strings. Constraining a family of conditional models in this manner is typically much more manageable computationally than working with a single constrained joint distribution. In addition, the use of conditional models is desirable for applications which process the input in a left-to-right fashion.

## 3. Iterative Scaling

The generalized iterative scaling algorithm of Darroch and Ratcliff [2] is one method for calculating the maximum entropy distribution (1). This algorithm assumes that the features $f_\alpha(h, w)$ are non-negative and sum to a constant, independent of $h$ and $w$:

$$M(h, w) \equiv \sum_\alpha f_\alpha(h, w) = M, \quad \text{for all } h, w. \tag{2}$$

Given these restrictions, the Darroch-Ratcliff algorithm begins with an initial model, typically the uniform distribution obtained by setting $\lambda_\alpha = 0$. In the iterative step, when the current model is $p_\lambda(w \,|\, h)$, the algorithm increments each parameter $\lambda_\alpha$ by an amount $\Delta\lambda_\alpha$ determined by

$$\Delta\lambda_\alpha = \frac{1}{M} \log\left( \frac{\sum_{h,w} \tilde{p}(h,w)\, f_\alpha(h,w)}{\sum_{h,w} \tilde{p}(h)\, p_\lambda(w\,|\,h)\, f_\alpha(h,w)} \right)\ .$$

Letting $\Delta\beta_\alpha = e^{\Delta\lambda_\alpha}$, we can express this update as choosing $\Delta\beta_\alpha$ to be the unique solution of the equation

$$\tilde{p}_\lambda[f_\alpha \Delta\beta_\alpha^M] = \tilde{p}[f_\alpha] \tag{3}$$

where $q[\,\cdot\,]$ denotes expectation with respect to $q$ and we use $\tilde{p}_\lambda$ to denote the distribution $\tilde{p}_\lambda(h,w) = \tilde{p}(h)\, p_\lambda(w\,|\,h)$.

While the restriction (2) on $M$ can always be enforced by introducing a "slack variable," it can be inconvenient to do so for conditional maximum entropy language models that typically have hundreds of thousands of features. In [1] an algorithm was introduced that extends the Darroch-Ratcliff procedure by relaxing the assumption that $M(h,w)$ is a constant. The updates for the improved algorithm are again given by equation (3), but with $M$ now interpreted as a random variable. When (2) holds, the algorithms are identical. In general, the algorithm which allows $M$ to vary is more natural and easier to implement. It also converges more quickly, by effectively increasing the step size taken toward the maximum entropy solution at each iteration.

## 4. Cluster Expansions

### 4.1. THE MAYER EXPANSION FOR A CLASSICAL GAS

If the Hamiltonian for a classical $N$-particle system is given by $H = \frac{1}{2}\sum_i p_i^2 + \sum_{i<j} v_{ij}$ and the system occupies a volume $V$, then the classical partition function of the system at temperature $T$ is given by

$$Q_N(V,T) = \frac{1}{h^{3N} N!} \int_{\mathbf{R}^{3N}} \int_V dp\, dq\, \exp\left( -\tfrac{1}{2}\beta \sum_i p_i^2 - \beta \sum_{i<j} v_{ij} \right)$$

where $\beta = 1/kT$ and $h$ is a constant introduced to make $Q_N$ dimensionless. Computing the integral over the momenta reduces this to

$$Q_N(V,T) = \frac{1}{\lambda^{3N} N!} \int_V dq\, \exp\left( -\beta \sum_{i<j} v_{ij} \right) \equiv \frac{1}{\lambda^{3N} N!} Z_N(V,T)$$

where $\lambda = \sqrt{2\pi\hbar^2/kT}$. The idea of the cluster expansion is to make a change of variables

$$\phi_{ij} = e^{-\beta v_{ij}} - 1$$

and expand $Z_N$ as a sum of products of $\phi_{ij}$:

$$Z_N(V,T) = \int_V dq \prod_{i<j}(1+\phi_{ij}) = \int_V dq\, \left( 1 + \sum_{i<j}\phi_{ij} + \sum_{i<j}\sum_{k<l}\phi_{ij}\phi_{kl} + \cdots \right)\ .$$

A convenient way to think about the integrals that need to be computed comes from expressing the various terms as graphs. If $N = 3$, for example, the integrands are represented as graphs as follows:

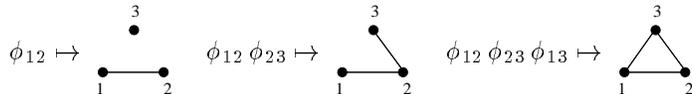

In terms of this correspondence, $Z_N = \sum_G S(G)$, where the sum is over all $N$-particle graphs and $S(G)$ is the appropriate integral; for example,

$$S( \quad ) = \int_V dq\, \phi_{12}\, \phi_{13}\,.$$

If a graph $G$ is disconnected, then $S(G)$ factors into a product of terms, and each connected component is referred to as a *cluster*. The *Mayer cluster integral* $b_l$ is given by $b_l = 1/l! \sum_{l\text{-}clusters\ G_l} S(G_l)$. Thus,

$$b_3 = \frac{1}{3!} S \left( \quad + \quad + \quad + \quad \right).$$

Simple combinatorial arguments lead to an expression for $Z_N$ in terms of the integrals $b_l$. While this is then carried further to obtain a series expansion for the grand partition function, our use of the method will simply make use of the discrete analogues of the integrals $b_l$ for conditional models. For more details on the statistical physics calculations we refer to [9].

## 4.2. CLUSTER EXPANSIONS FOR CONDITIONAL MAXENT MODELS

The computation necessary to carry out the iterative scaling algorithm described in Section 3 is naturally divided into two parts. First, for a conditional maximum entropy model of the form (1), it is necessary to compute the partition functions $Z_\lambda(h)$ for each history $h$ such that $\tilde{p}(h) > 0$. Using the notation from statistical physics, we make the change of variables $\phi_\alpha(h, w) = e^{\lambda_\alpha f_\alpha(h,w)} - 1$ so that $Z_\lambda(h)$ can be expressed as

$$Z_\lambda(h) = \sum_w \prod_\alpha (1 + \phi_\alpha) = \sum_w \left( 1 + \sum_\alpha \phi_\alpha + \sum_{\alpha, \alpha'} \phi_\alpha\, \phi_{\alpha'} + \cdots \right).$$

In analogy with the classical expansion, this expresses the normalization $Z_\lambda(h)$ as a sum of *cluster integrals*, where $\sum_w \sum_\alpha \phi_\alpha(h, w)$ is the order one cluster, $\sum_w \sum_{\alpha, \alpha'} \phi_\alpha\, \phi_{\alpha'}$ is the order two cluster, and the highest order cluster that needs to be computed is the order-$M$ cluster where $M$ is the largest value of $\sum_\alpha f_\alpha(h, w)$.

This gives an *exact* expression for $Z_\lambda(h)$ as a telescoping sum. The point of using this technique, as we will explain further in the following section, is that computation of the individual clusters can be significantly more efficient than computing $Z_\lambda(h)$ directly. Furthermore, the computation of the clusters can be shared across different histories. The use of Cheeseman's method [10,11] of reordering summations within a cluster can provide further savings.

The second computation that is necessary is the calculation of the coefficients of $\Delta\beta_\alpha$ in the expectation $\tilde{p}_\lambda[f_\alpha\Delta\beta_\alpha^M]$ that appears in the scaling equation (3). In a manner similar to that described above, we expand in terms of $\phi_\gamma$ to obtain

$$\tilde{p}_\lambda[f_\alpha\Delta\beta_\alpha^M] = \sum_h \frac{\tilde{p}(h)}{Z_\lambda(h)} \sum_w \left(1 + \sum_\gamma \phi_\gamma + \sum_{\gamma,\gamma'} \phi_\gamma\,\phi_{\gamma'} + \cdots\right) f_\alpha(h,w)\,\Delta\beta_\alpha^{M(h,w)}\,.$$

Here again, indirect computation of the coefficients through the calculation of the individual cluster terms can be significantly more efficient than direct computation.

The primary savings that this technique affords results from its avoidance of an explicit summation over the entire target vocabulary for each history. In addition, it can make hashing for feature lookup unnecessary. While we do not generally obtain better theoretical computational complexity, this simple trick can result in substantial savings in the computation necessary for carrying out generalized iterative scaling. We will now give further details of these calculations for a simple topic-dependent bigram model developed for use in a speech recognition system.

## 5. Example: A Topic-Dependent Language Model

In this section we describe the application of the cluster expansion to the training of a topic-dependent bigram model of the *Switchboard* corpus [12] for use in a speech recognition system. This corpus comprises approximately three million words of text, transcribed from more than 150 hours of speech collected from telephone conversations. An important aspect of the Switchboard corpus is that the conversations are restricted to 70 different topics. To take advantage of this structure, we trained a maximum entropy language model whose constraints were of three types. In addition to unigram and bigram constraints, we introduced topic-dependent unigram constraints for those words having the greatest mutual information with the topic.

More precisely, the model that we constructed was specified as follows. Conditioning on a word history $h$ which ends in a word $w'$, the probability of predicting $w$ is given by

$$p(w\,|\,h) = \sum_t p(\text{topic} = t\,|\,h)\,p(w\,|\,h,t) = \sum_t p(\text{topic} = t\,|\,h)\,p(w\,|\,w',t)\,.$$

This model has two components: a *topic prediction* model $p(\text{topic} = t\,|\,h)$ and a *word prediction* model $p(w_j\,|\,t,w_i)$. (The topic prediction model is not discussed here.) The word prediction model is constructed as a conditional maximum entropy distribution of the form

$$p_\lambda(w_j\,|\,t,w_i) = \frac{1}{Z_\lambda(i,t)}\exp\left(\lambda_{ij} + \lambda_i + \lambda_{tj}\right)\,.$$

We thus place constraints on the model so that it agrees with the bigram and unigram frequencies as they appear in the data. In addition, we constrain the topic-dependent unigrams, corresponding to the parameters $\lambda_{tj}$, for those words $w_j$ that appear with sufficiently high mutual information with topic $t$. For example, the topic-independent bigram constraint equations take the form

$$\sum_t \tilde{p}(w_i, t)\, p(w_j \mid w_i, t) = \tilde{p}(w_i, w_j) \equiv c_{ij}$$

where $\tilde{p}$ is the empirical distribution, and the corresponding scaling equations update $\lambda_{ij}$ by an amount $\Delta\lambda_{ij} = \log\Delta\beta_{ij}$, where $\Delta\beta_{ij}$ is the unique positive solution to the equation

$$\sum_t \tilde{p}(w_i, t)\, p_\lambda(w_j \mid w_i, t)\, \Delta\beta_{ij}^{M(i,t,j)} = c_{ij} \,.$$

The constraint and scaling equations for the parameters $\lambda_i$ and $\lambda_{tj}$ are similar.

To apply the cluster expansion technique to this model we express the partition functions $Z_\lambda(i, t)$ in terms of the variables $\phi_\alpha = e_\alpha^\lambda - 1$ and expand $Z_\lambda(i, t) = \sum_j (1 + \phi_j)(1 + \phi_{tj})(1 + \phi_{ij})$ into a sum of four cluster "integrals"

$$Z_\lambda(i, t) = b_0 + b_1(i, t) + b_2(i, t) + b_3(i, t) \,.$$

Using a variant of the physicists' graph notation that is appropriate for conditional models, we can express these terms as a sum over all configurations of a set of graphs; for example,

$$b_2(i, t) = S\left( \begin{array}{c} \varepsilon \bullet \\ t \bullet \\ i \bullet \end{array} + \begin{array}{c} \varepsilon \bullet \\ t \bullet \\ i \bullet \end{array} + \begin{array}{c} \varepsilon \bullet \\ t \bullet \\ i \bullet \end{array} \right) \,.$$

In these figures the unlabeled vertex is summed over, and an edge connecting the vertex labeled $\varepsilon$ denotes a unigram term $\phi_j$. Thus,

$$b_3(i, t) = S\left( \begin{array}{c} \varepsilon \bullet \\ t \bullet \\ i \bullet \end{array} \right) = \sum_j \phi_j\, \phi_{tj}\, \phi_{ij} \,.$$

We use the fact that $\phi_\alpha = 0$ unless $\lambda_\alpha$ is a parameter that is being estimated. This is what allows the above telescoping summation to be carried out efficiently; for example, the summation $\sum_j \phi_j \phi_{ij}$ is carried out only over those indices $j$ for which the bigram $(w_i, w_j)$ is constrained. The largest cluster, $b_3(i, t)$, involves a summation over all those indices $j$ for which the bigram $(w_i, w_j)$ is constrained *and* $w_j$ is a topic word for topic $t$. The cluster integrals for the various values of $(i, t)$ with $\tilde{p}(w_i, t) > 0$ can be calculated simultaneously by a single pass through appropriately constructed data structures, and require no expensive hashing of the bigram parameters. A very similar analysis is applied to the task of computing the coefficients of the iterative scaling equations for all of the parameters. When we implemented this technique for the topic-dependent model, the resulting calculation was more than 200 times faster than the direct implementation of the iterative scaling algorithm.

## 6. Summary

Our use of the cluster expansion for the language model presented in Section 5 demonstrates that this technique can be an important tool for reducing the computational burden of computing maximum entropy language models. The method also applies to higher order models such as "trigger models" [8], where occurrences of words far back in the history can influence predictions by the use of long-distance bigram parameters. As a general technique, however, the method is limited in its usefulness. As in statistical mechanics, when the number of interacting constraints is large (*i.e.*, when the gas is dense), the cluster expansion is of little use in computing the exact maximum entropy solution. For such cases the use of approximation techniques should be investigated.